\documentclass[conference,compsoc]{IEEEtran}
\ifCLASSOPTIONcompsoc
  \usepackage[nocompress]{cite}
\else
  \usepackage{cite}
\fi
\ifCLASSINFOpdf
\else
\fi
\usepackage{graphicx}
\usepackage{algorithm}
\usepackage{algorithmic}
\usepackage{amsmath}
\usepackage{amssymb}
\usepackage{hyperref}
\DeclareMathOperator*{\argmin}{argmin}

\usepackage{multirow}
\usepackage{caption}
\usepackage{subfigure}
\usepackage[round]{natbib}

\begin{document}
%
\title{Manipulating Predictions over Discrete Inputs in Machine Teaching}

\author{
\IEEEauthorblockN{Xiaodong Wu}
\IEEEauthorblockA{Queen's University
at Kingston, Canada \\
xiaodong.wu@queensu.ca}
\and
\IEEEauthorblockN{Yufei Han}
\IEEEauthorblockA{INRIA, France\\
yfhan.hust@gmail.com}
\and
\IEEEauthorblockN{Hayssam Dahrouj}
\IEEEauthorblockA{University of Sharjah, UAE\\
hayssam.dahrouj@gmail.com}
\and

\IEEEauthorblockN{$~~~~~~~~~~~$Jianbing Ni}
\IEEEauthorblockA{$~~~~~~~~~~~$Queen's University
at Kingston, Canada\\
$~~~~~~~~~~~$jianbing.ni@queensu.ca}
\and
\IEEEauthorblockN{$~~~~~~~~$Zhenwen Liang}
\IEEEauthorblockA{$~~~~~~~~$Notre Dame, America\\
$~~~~~~~~$zliang6@nd.edu}
\and
\IEEEauthorblockN{$~~~~~~~~~~~$Xiangliang Zhang}
\IEEEauthorblockA{$~~~~~~~~~~~$Notre Dame, America\\
$~~~~~~~~~~~$xzhang33@nd.edu}
}


%


\maketitle

\begin{abstract}
Machine teaching often involves the creation of an optimal (typically minimal) dataset to help a model (referred to as the `student') achieve specific goals given by a teacher. 
While abundant in the continuous domain, the studies on the effectiveness of machine teaching in the discrete domain are relatively limited. This paper focuses on machine teaching in the discrete domain, specifically on manipulating student models' predictions based on the goals of teachers via changing the training data efficiently. We formulate this task as a combinatorial optimization problem and solve it by proposing an iterative searching algorithm. Our algorithm demonstrates significant numerical merit in the scenarios where a teacher attempts at correcting erroneous predictions to improve the student's models, or maliciously manipulating the model to misclassify some specific samples to the target class aligned with his personal profits. Experimental results show that our proposed algorithm can have superior performance in effectively and efficiently manipulating the predictions of the model, surpassing conventional baselines.
\end{abstract}


%
\IEEEpeerreviewmaketitle

\section{Introduction}

Machine teaching is a learning paradigm that involves the construction of an optimal dataset by a teacher model, which enables a student model to learn and achieve the teacher's desired objectives. It can be deemed as an inverse problem to machine learning, which is first proposed by \cite{goldman1995complexity}. Due to its effective application in dataset organization, it has attracted great interests from the academia. Many previous work \citep{liu2016teaching,zhu2013machine} study the teaching dimension in the machine teaching problem, which is the size of the smallest organized dataset for students to learn a valid model. In addition to the conventional approach of having a student model mimic the behavior of a teacher model with small datasets, machine teaching can have an alternative setting where the prediction of a student model is manipulated by the teacher model. 

In the beneficial setting, the teacher leverages the constructed dataset to guide the student model to behave in a more accurate or robust manner \citep{han2020robust}. For example, some technology companies may seek assistance from domain experts to help improve their poorly behaved machine learning models. Teachers in such setting acts like DUTI (Debugging Using Trusted Items). In another example, \cite{zhang2018training} propose to change training labels as few as possible by relax the improvement problem into a continuous optimization problem. 
Conversely, the malicious setting resembles a targeted poisoning attack and has recently garnered noticeable attention \citep{alfeld2016data,shafahi2018poison}. 
Some prior works \citep{aghakhani2021bullseye} apply feature collision attacks to tamper the images 
so that the predictions of these images can be manipulated. 
For both settings, there are already abundant research works in continuous domain, e.g., images, because the continuous features (pixels) therein can be modified directly based on the same continuous gradients, while few attention has been paid to the discrete domain, which is a critical problem, considering security risks of current powerful nature language models. 

We specifically investigate machine teaching in the context of systems that rely on discrete inputs in both setting, which are prevalent in machine learning enabled systems, such as the clinical prediction systems built on Electronic Health Records (EHR) \citep{ma2017}.
Despite the importance of this field, it remains relatively unexplored, primarily due to the inherent challenges associated with discrete inputs. Firstly, machine teaching in the discrete domain can be formulated as a combinatorial optimization problem, which has exponentially expanding search space, resulting in the difficulty in constructing a proper dataset. Secondly, the existing methods in the continuous domain which use gradient information of the target model to guide data poisoning cannot be directly applied on discrete training data. This limitation arises from the fact that the new inputs calculated from the continuous gradient vectors and original discrete inputs should remain discrete.

Furthermore, the selection of base samples in continuous domain can be done randomly, but in discrete domain, a deliberate and meticulous approach is imperative. Specifically, the samples from the same class in continuous domain often have similar features, thereby enabling smooth manipulation to construct an effective teacher dataset. However, applying the same level of perturbation on different categorical variables can lead to dramatically different prediction results, as the importance of categorical variables varies. Therefore, 
selecting appropriate base samples and determining which features to perturb become a crucial challenge. 
A related study by \cite{wang2020attackability} propose an optimization-based selection algorithm to solve the feature selection problem in an evasion attack scenario, where the trained model remains unchanged. This algorithm tackles the challenge of identifying the most influential features to perturb for achieving the desired attack objectives. 

Our objective is to construct a compact teacher dataset that, when added to the training data, can effectively manipulate the predication of the student model, either correcting erroneous predictions, or leading the model to make incorrect predictions. The key challenge is to efficiently find a minimized teacher dataset that achieves the desired manipulation. 
To address this challenge, we propose an iterative algorithm called \textbf{D}iscrete \textbf{M}achine \textbf{T}eaching (DMT) that formulates the teacher dataset construction as a combinatorial optimization problem and tries to solve it with two steps.
\textbf{1) Teacher dataset construction}. We initially select instances from the training dataset to form the base of the teacher dataset. For each instance, we determine modifications to be made on categorical attributes. This involves adding new features, removing original features, or replacing the original ones with new ones; and \textbf{2) student model update. } We combine the teacher dataset with the current training data and retrain the model. The new model is evaluated to examine if the manipulation target is reached. If not, we return to step 1) and repeat. 
As demonstrated in \citep{liu2017iterative}, this iterative machine teaching approach can generate smaller teacher dataset and facilitates faster convergence. The reason is that this particular approach allows the teacher to communicate with the student model in multiple rounds. In each round, the teacher can observe the status of the student and intelligently choose the most suitable samples to guide the training process towards the desired target. The key innovation of our DMT lies in the first step. We design  effective score functions for selecting the base samples and features to be perturbed. Based on the score functions, we achieve successful manipulations while significantly reducing the searching space. {Our code is shared in Appendix 1.}

The main contributions of our work can be summarized as follows.
\begin{itemize}
\item We propose to solve the machine teaching problem with discrete categorical inputs, which is still an under-explored problem. 

\item We design an efficient method for constructing the teacher dataset, which estimates the impact of individual samples and their categorical features to select the most influential samples and features for perturbation.    
We incorporate the teacher dataset in an iterative machine teaching framework, enabling the effective manipulation of student model prediction.  

\item We evaluate our DMT method on three discrete datasets and observe successful manipulations on the predictions. Furthermore, our evaluations reveal that our method outperforms the baseline methods in terms of efficiency.
\end{itemize}

In the remainder of this paper, we first discuss the related works and their limitations. Then, we define our problem by providing the problem formulation. Next, we discuss the details of our methodology. Finally, we evaluate our method on three discrete datasets and analyze the results.

\section{Related Work}

This section reviews the related work about machine teaching and how to improve or tamper the predictions in machine teaching.

\subsection{Machine Teaching}
Machine teaching, a new machine learning paradigm, was proposed by \cite{goldman1995complexity}. It aims at creating an optimal dataset from which the student model learns to meet the goal of the teacher. 
Subsequently, fundamental definitions and concepts of machine teaching were further developed in \citep{zhu2015machine}.
The current focus of machine teaching study lies in the teaching dimension, which refers to the size of the minimal dataset required to teach students to attain the desired objectives \citep{chen2018understanding,kane2017active}. 
It is desired to have a minimized teaching dimension while achieving the training target. Previous researches found that introducing interactions between teacher and student model can effectively reduce the teaching dimension. 
For example, \cite{liu2017iterative} showed how iterative interactions between the teacher and the student can enhance the process.
The iterative approach significantly reduces the teaching dimension and accelerates convergence for students. Typically, this method is applied in the continuous domain, where the optimization problem can be solved with analytical expressions. However, when used in the discrete domain, the optimization problem cannot satisfy KKT conditions and generally becomes an NP-hard problem. 
Therefore, we propose an efficient iterative machine teaching method that incorporates an efficient searching strategy to craft an optimal dataset for discrete data.

\subsection{Prediction Manipulation}
According to the goals of teachers, teachers' impact on student models can be classified into two types: prediction improvement and prediction tampering. In the first case, teacher behaves like DUTI, aiming to improve the classification prediction results by modifying training items. 
This debugging approach has been extensively studied with machine learning techniques \citep{cadamuro2016debugging,bhadra2015correction}. For example, 
\cite{han2020robust} designed a collaborative machine teaching method that solves the problem by tuning the training samples with a few trusted items. They optimize the training samples by utilizing the gradient information from correct classification, resulting in significant improvement of the student models' performance.

In the second case, the teacher's objective is similar to a poisoning attack where the teacher manipulates the student models to do wrong predictions by tampering with model's prediction. 
Studies in this field are abundant in the continuous domain. However, in discrete domains, existing works primarily focus on solving backdoor attack problem, a special type of poisoning attack \citep{guo2022overview,salem2022dynamic}. The goal of the backdoor attack is to establish a new relation between a specific target label and some pre-designed triggers during training. This relationship leads to test samples containing the pre-designed triggers being classified as the target class. For example, 
\cite{schuster2020humpty} and \cite{yang2021careful} explore controlling the word embeddings within target models.
Besides, some other works aim to make their attacks stealthier and harder to defend against. For instance, \cite{wallace2020concealed} propose to using poisoning sentences that do not contain trigger phrases. This approach achieves high attack success rate and is very difficult to detect because of the hidden trigger words. 
 
All the above methods are either in the continuous domain or in the discrete domain but requiring access to tamper both the training and testing data (backdoor attack).
In the general setting, the focus is solely on creating a perturbed training dataset. Therefore, there is a lack of exploration in discrete domains where new relation between features and target can be directly established without relying on triggers.
Considering the vast applications of machine teaching in discrete domain, manipulating student model's performance in various NLP tasks is a meaningful but challenging problem.
Our work focuses on generating optimal datasets for efficient machine teaching in the discrete domain. 

\section{Methodology}

In this section, we formulate our problem of teacher dataset construction and propose the discrete machine teaching methodology to manipulate the performance of the student models.

\subsection{Problem Formulation}
Let $D_{clean}$ be a clean training dataset without any perturbation, and contain a set of training data samples $\{x_i, y_i\}$, where $y_i$ is the label of $x_i$. A classifier $f_\theta$ is learned from $D_{clean}$ by minimizing the objective function $\mathcal{L}(D_{clean},\theta) = \sum_{(x_i, y_i) \in D_{clean}} l(f_\theta(x_i),y_i)$, where $l(.)$ is a typical classification loss, e.g., cross entropy.  The obtained optimal parameter $\theta_{clean}^*$ is expected to make correct predictions in the testing dataset. 

The teacher aims to develop a teacher dataset $D_{teacher}$ such that the learned $\theta^*$ can manipulate prediction on target samples, i.e., 
$f_{\theta^*}(x_j) =\tilde{y}_j$, where  $\tilde{y}_j$ is the target label that the teacher would like $f_{\theta^*}$ to produce with the input of the target sample $x_j$. In addition, the teacher dimension (the size of $D_{teacher}$) should be minimal. In our discrete setting, $D_{teacher}$ is constructed by combining the training data $D_{clean}$ with the dataset $D_{perturbed}$ perturbed from $D_{base}$, which is a selected subset of $D_{clean}$. 
Therefore, the overall discrete machine teaching problem is defined by the objective function as follows, 

\begin{equation}\label{eq:equation1}
\begin{gathered}
    \mathop{\min}_{D_{perturbed} } \mathcal{L}(D_{target};\mathop{\arg\min}_{\theta} \mathcal{L}(D_{clean} \cup D_{perturbed};\theta)) \\
s.t.    \quad    \text{diff}(x_{i},  \hat{x}_{i}) \leq \epsilon, \;  x_{i} \in D_{base}, \hat{x}_{i} \in D_{perturbed}, 
\end{gathered}
\end{equation}
where $\text{diff}(x_{i},  \hat{x}_{i} ) \leq \epsilon $ establishes the perturbation budget, $D_{target}$ is the set of targeted samples and labels.
Suppose a discrete instance $x$ is represented by transforming categorical features into binary variables, i.e., $x \in \{0,1\}^{M\times N}$, where $M$ is the number of categorical features and $N$ is the number nominal values that each feature has. A value 1 in $x$ indicates the presence of the value within the categorical feature, and 0 means absence. The perturbed $\hat{x}$ can be obtained by changing 1 to 0 in $x$ (\emph{deletion}), or changing 0 to 1 in $x$ (\emph{insertion}). It is also possible to make several changes and have \emph{substitution} in $x$. When $M\times N$ is large, it would be expensive to take exhaustive search for finding the minimal set of changes that optimizes Eq. (1).

\subsection{Teacher's Capability}
The goal of teachers in machine teaching is to shape student model's behavior, either by improving their performance in specific tasks, or by stealthily tampering students' predictions to benefit teachers. To achieve this goal, the teachers first should have the necessary computational resources and time construct their teacher datasets and possession the following knowledge: 1) The information about the training data used to train the student model. 2) The architecture of the student model. 3) The details of the target samples and their expected predictions. Secondly, they should have the access to the training data containing specific target samples that they aim to influence. Thirdly, teachers can modify a portion of the training data or insert generated samples into the dataset, but adhering to a manipulative budget $\epsilon$. However, they do not have the capacity to modify the test data. Fourthly, teachers can obtain gradient vectors or middle outputs of the student model through querying, which guides their modification process. We next introduce our solution for discrete machine teaching, which includes the finding of $D_{base}$, and the efficient constructions of $D_{perturbed}$ and $D_{teacher}$. 

\subsection{DMT Method}
\renewcommand{\algorithmicrequire}{\textbf{Input:}}
\renewcommand{\algorithmicensure}{\textbf{Output:}}

\begin{algorithm}[t]
\caption{Discrete Machine Teaching}
\begin{flushleft}
\textbf{Input}: Original clean dataset $D_{clean}$, target sample $x_j$, the corresponding target class $\tilde{y}_j$, a student model $f_{\theta}(.)$, the max number of iteration $T$\\
\textbf{Output}: Manipulated model $f_{\theta^*}(.)$
\end{flushleft}
\begin{algorithmic}[1]
\STATE $\theta^* = \argmin\limits_{\theta} \mathcal{L}(D_{clean};\theta)$;
\STATE $D_0 = D_{clean}$;
\FOR{$t=1$ to $T$}
\STATE $D_{base} = k$-nn$(x_j, D_{t-1})$;
\STATE $D_{perturbed} = GGGM(D_{base}, x_j, \theta^*)$;
\STATE $D_t = Combine(D_{t-1}, D_{perturbed})$;
\STATE $\theta^* = \argmin\limits_{\theta} \mathcal{L}(D_t;\theta)$;
\IF {$f_{\theta^*}(x_j) = \tilde{y}_j$}
    \STATE \textbf{break}
\ENDIF
\ENDFOR
\end{algorithmic}
\label{alg:DMT}
\end{algorithm}

The proposed DMT method is presented in Algorithm \ref{alg:DMT}. It is an iterative algorithm including basic steps of \textbf{1) teacher dataset construction} (line 4-6), and \textbf{2) student model update} (line 7).   The new model is tested if the manipulation target is reached (line 8). If not, go to step 1) and repeat until hitting the maximum number of iterations $T$. The construction of the teacher dataset consists of three steps below.

\vspace{+0.2cm}
\noindent
\textbf{Selection of $D_{base}$.}
To minimize $D_{teacher}$, $D_{base}$ should consist of the most influential base samples that have the capability to alter the prediction of target sample $x_j$.
Intuitively, the samples in $D_{base}$ should surround $x_j$. 
We select base samples by determining the $k$-nearest neighbors of the target sample $x_j$ in the training dataset. Since the data have discrete categorical features, we use the Jaccard distance for the nearest neighbor identifying. 
Initially, $D_{base}$ is selected from the clean dataset $D_0 = D_{clean}$. When the teacher dataset $D_t$ is updated, the selection of $D_{base}$ in next round would have access to the updated $D_t$, including the perturbed dataset $D_{perturbed}$ at step $t$. The selected $k$ samples would be used to create $D_{perturbed}$ based on Gradient Guided Greedy Method (GGGM), which is introduced in the next section. In the iterative teaching process, $k$ is also named as \emph{step size}. Our evaluation results demonstrate the impact of $k$  on the teaching performance.

\vspace{+0.2cm}
\noindent
\textbf{Construction of $D_{perturbed}$.}
The construction of $D_{perturbed}$ is to select the most powerful features in each sample of $D_{base}$. The details are shown in Algorithm \ref{alg:GGGM}. 
For each sample in $D_{base}$, we use gradients to first filter out the candidate changes  that have the largest absolute  gradient values (line 5-6), as the gradient magnitude of $f_{\theta^*}(x)$  over $x$ implies the contribution to prediction \citep{lei2018discrete}. Unlike what was done in continuous domain, these gradients cannot be used directly to construct $D_{perturbed}$, but can serve as guiding which feature to be selected and modified. 
Then, we evaluate the impact of these candidate changes by designing a function $g$ and choosing those with strongest potential to manipulate the prediction of $x_j$  (line 7). The perturbations made to obtain $\hat{x}$ is controlled within the budget $\epsilon$.

To evaluate the impact of modifying a subset of candidate features, we design two score functions $g(\cdot)$. 
The first one is based on the Euclidean distance function $g_{dist}$, which measures the difference of predictions made on the target sample $x_j$ and the perturbed sample $\hat{x}$, 
\begin{equation}
\begin{gathered}
    g_{dist}( \hat{x} , x_j, \theta) = 
    \Vert f_{\theta^*}(x_j)  - f_{\theta^*}(\hat{x})    \Vert_2 .
\end{gathered}
\label{eq:equation5}
\end{equation}
A perturbation is preferred if the perturbed $\hat{x}$ has a closer prediction score compared with the score of the target sample $x_j$.

The second method is the alignment function $g_{align}$ inspired from \cite{geiping2020witches}:
\begin{equation} \small
\begin{gathered}
    g_{align}(\hat{x}, x, x_j, \theta^*) = - \lambda * align(\hat{x}, x_j, \theta^*) \\
    - (1-\lambda) * align(\hat{x}, x, \theta^*), 
\end{gathered}
\label{eq:equation4}
\end{equation}

where $\lambda$ is a trade-off parameter that combines two alignment scores: one measures how a perturbed sample $\hat{x}$ is aligned with the targeted sample $x_j$; the other one is measuring the alignment between the perturbed sample $\hat{x}$  and the clean sample $x$.
The alignment score is defined as the cosine distance of the gradient vectors of two given   samples, which is defined as:
\begin{equation} \small
        align(x_a, x_b, \theta) = \frac{\left<\bigtriangledown_{\theta}l(f_{\theta}(x_a), y_a),                 \bigtriangledown_{\theta}l(f_{\theta}(x_b),\tilde{y}_b)\right>} {\Vert\bigtriangledown_{\theta}l(f_{\theta}(x_a), y_a)\Vert   \Vert\bigtriangledown_{\theta}l(f_{\theta}(x_b),\tilde{y}_b)\Vert},
\end{equation}
The idea behind this design is to 
strike a balance between achieving the manipulation goal (via comparing $\hat{x}$ and $x_j$) and controlling the modification extend (via comparing $\hat{x}$ and clean $x$). 

\begin{algorithm}[t]
\caption{Gradient Guided Greedy Method (GGGM)}
\begin{flushleft}
\textbf{Input}:   base dataset $D_{base}$, a trained student model  $f_{\theta^*}(.)$, the target sample $x_j$, the manipulative budget $\epsilon$, the candidate set size $q$ \\
\textbf{Output}: perturbed dataset   $D_{perturbed}$
\end{flushleft}
\begin{algorithmic}[1]
\STATE $D_{perturbed} = \emptyset$
\FOR {$x$ in $D_{base}$}
\STATE $\hat{x}^1=x$
\FOR{$p=1$ to $\epsilon/q$}
\STATE $r = \bigtriangledown_x f_{\theta^*} (\hat{x}^p)$
\STATE $s =\{(m,n)\}_{|r_{mn}| \in \;  \text{top-}q\text{-values}(|r_{.,.}|)}$
\STATE $S = \argmin\limits_{l \subset s} g($modify$ (\hat{x}^{p},l), x, x_j, \theta^*)$
\STATE $\hat{x}^{p+1}$ = modify $(\hat{x}^{p},S)$
\ENDFOR
\STATE $D_{perturbed} \leftarrow D_{perturbed} \cup \hat{x}^p$
\ENDFOR
\end{algorithmic}
\label{alg:GGGM}
\end{algorithm}

\vspace{+0.2cm}
\noindent
\textbf{Combination of $D_{perturbed}$ and $D_{base}$.}
In Algorithm \ref{alg:DMT} line 6, after obtaining the new dataset $D_{perturbed}$, a strategy should be applied to combine this dataset with the base dataset $D_{t-1}$ from the last iteration. Here, we propose to incrementally add new generated fake data $D_{perturbed}$ to the base dataset $D_{t-1}$. Consequently, the size of the modified dataset $D_t$ increases by a step size $k$, i.e., the cardinality of $D_{perturbed}$, in each iteration. Due to the nature of this strategy, incremental addition allows the effect of the manipulation dataset in each iteration gradually manipulates the prediction of the target sample toward the target class step by step. Furthermore, by keeping the step size small, the distribution of $D_{t-1}$ would not be changed dramatically, enabling the model to predict other test samples with a reasonable accuracy. An alternative strategy is to \emph{replace} the base samples in $D_{t-1}$ that correspond to the samples in $D_{perturbed}$.
However, as this strategy creates larger changes in $D_{t-1}$, the distribution of $D_{t-1}$ would be significantly affected. Further comparisons between these two strategies, i.e., \emph{addition} and \emph{replacement}, are provided in Appendix 4.

\vspace{+0.2cm}
\noindent

In addition, when there are multiple samples in $D_{target}$, and the teacher aims to manipulate a group of targeted samples, we calculate the average distance regarding all target samples in the step of base sample selection (line 4 of Algorithm \ref{alg:DMT}).  
Similarly, for line 7 of Algorithm \ref{alg:GGGM}, $g_{dist}$ is calculated by taking the average over all target samples. 
When using the alignment function in Eq.(3), we first average the gradients across all target samples, and then use this averaged gradient to calculate the alignment score.
\begin{figure}[t!]
\vspace{.3in}
\centering
\includegraphics[width=0.45\textwidth]{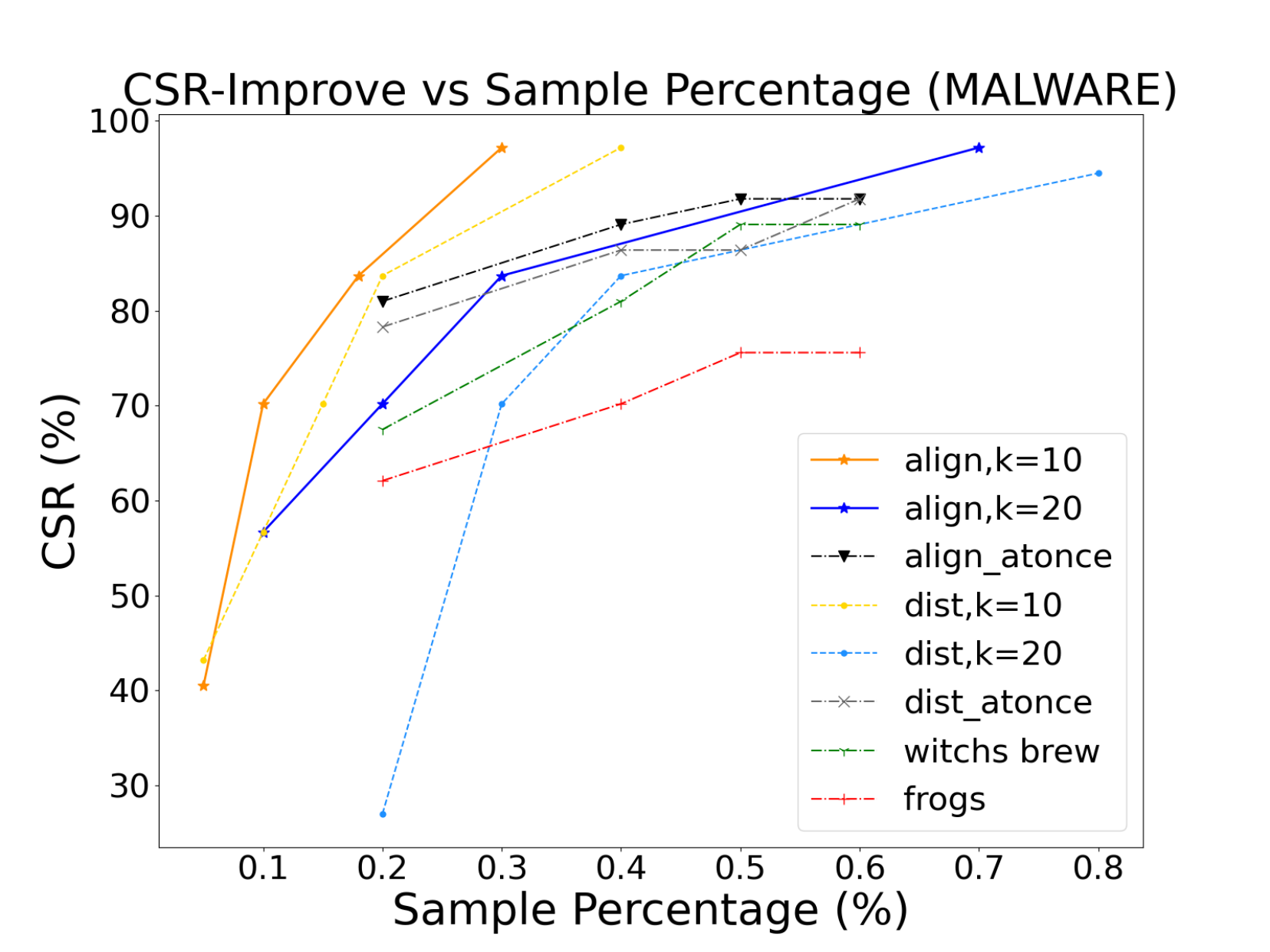}\\
\includegraphics[width=0.45\textwidth]{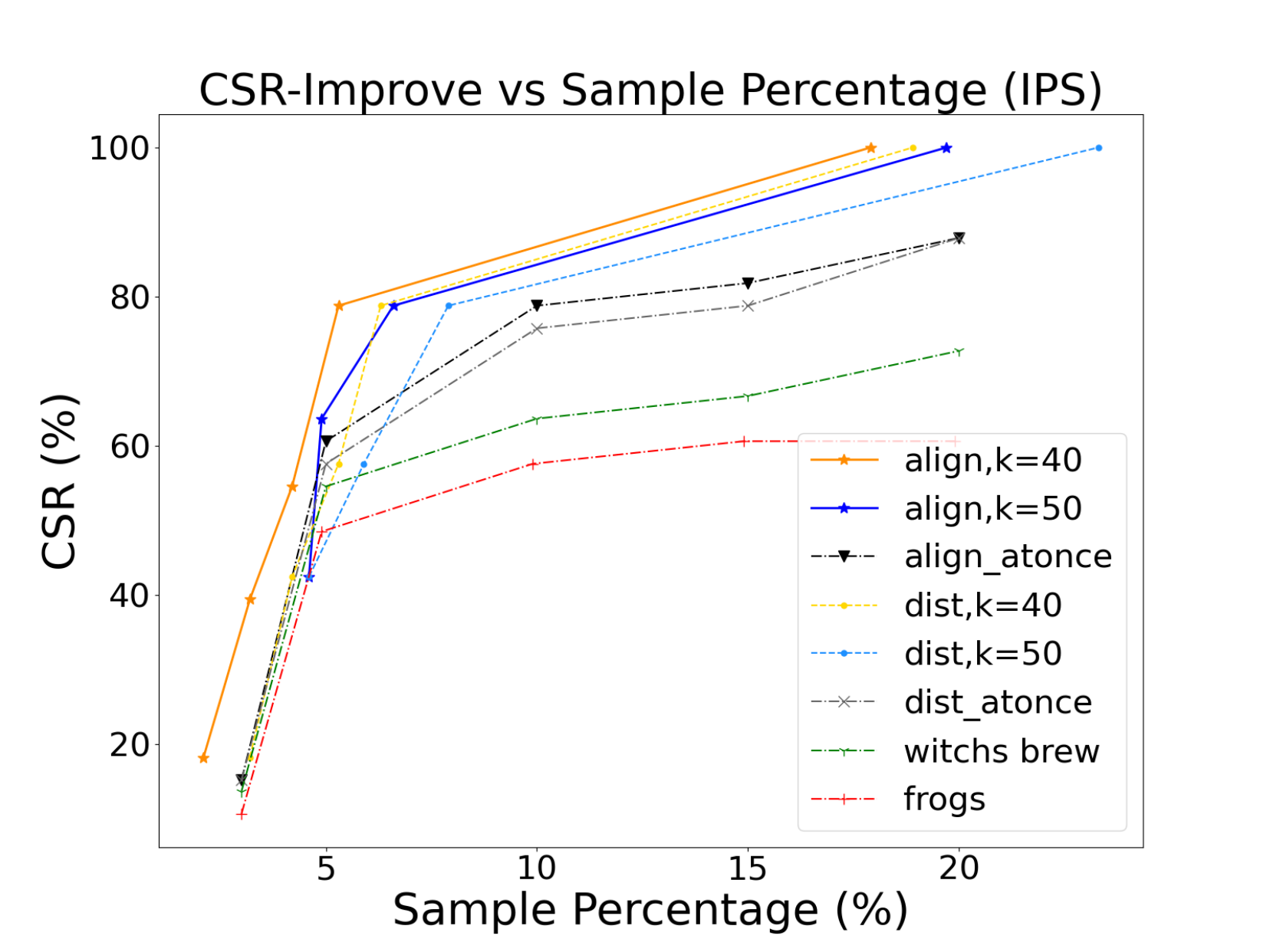}
\vspace{.3in}
\caption{
The CSR of Different Methods on the Performance Improvement Task When Varying the Percentages of Samples Allowed to Change.}
\label{fig:malware_improve}
\end{figure}

\begin{figure}[t!]
\vspace{.3in}
\centering
\includegraphics[width=0.45\textwidth]{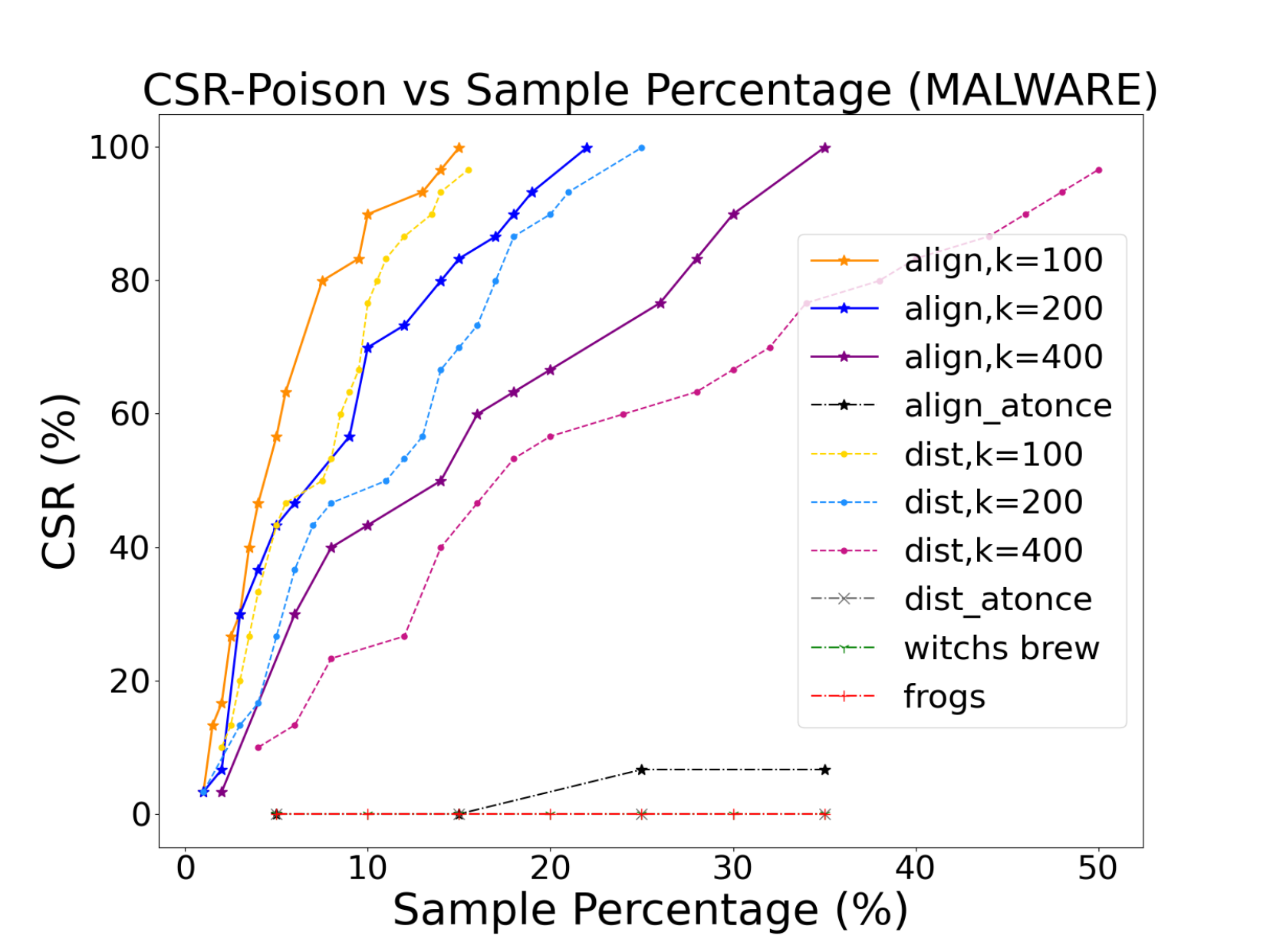}\\
\includegraphics[width=0.45\textwidth]{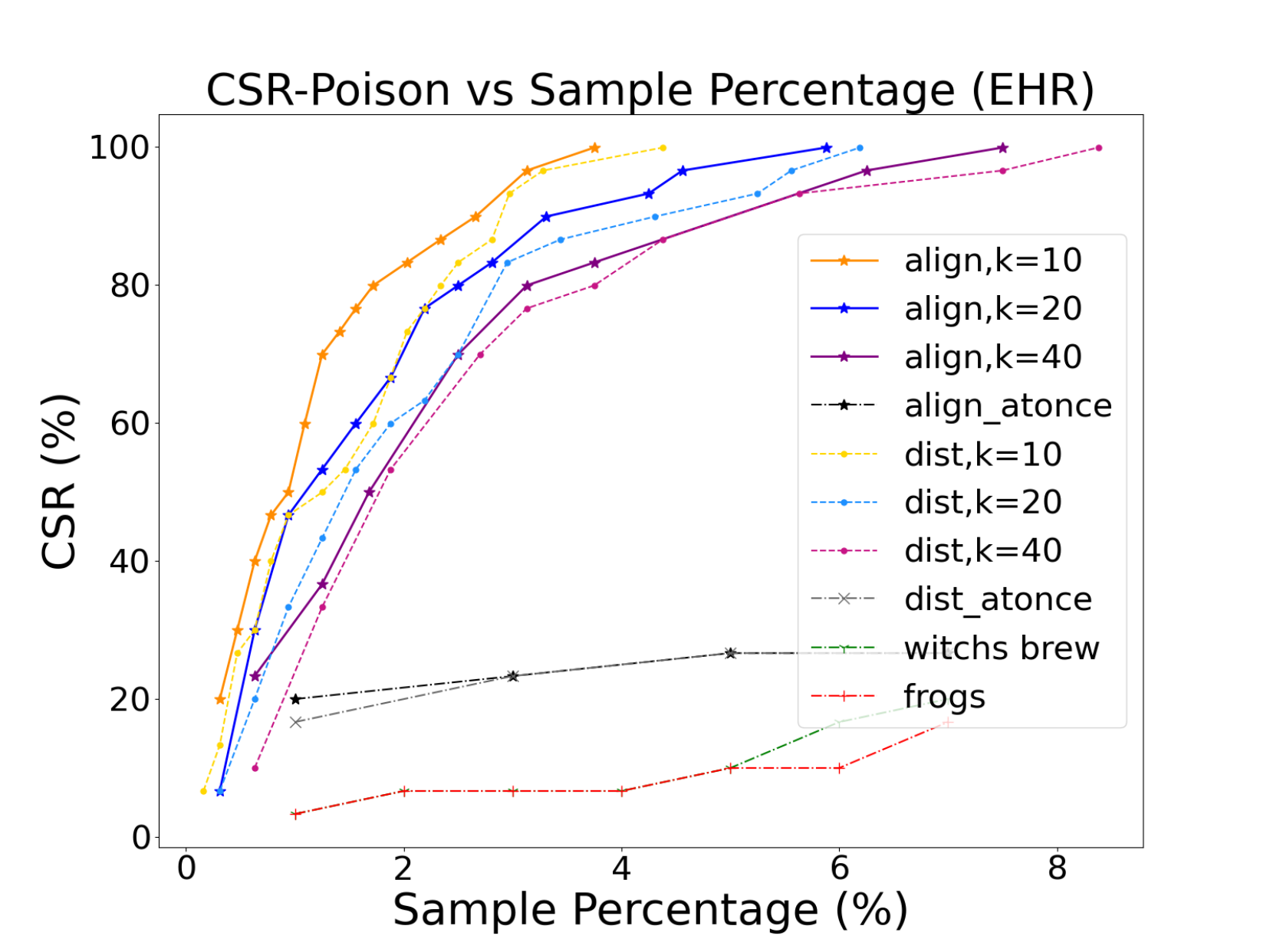}\\
\includegraphics[width=0.45\textwidth]{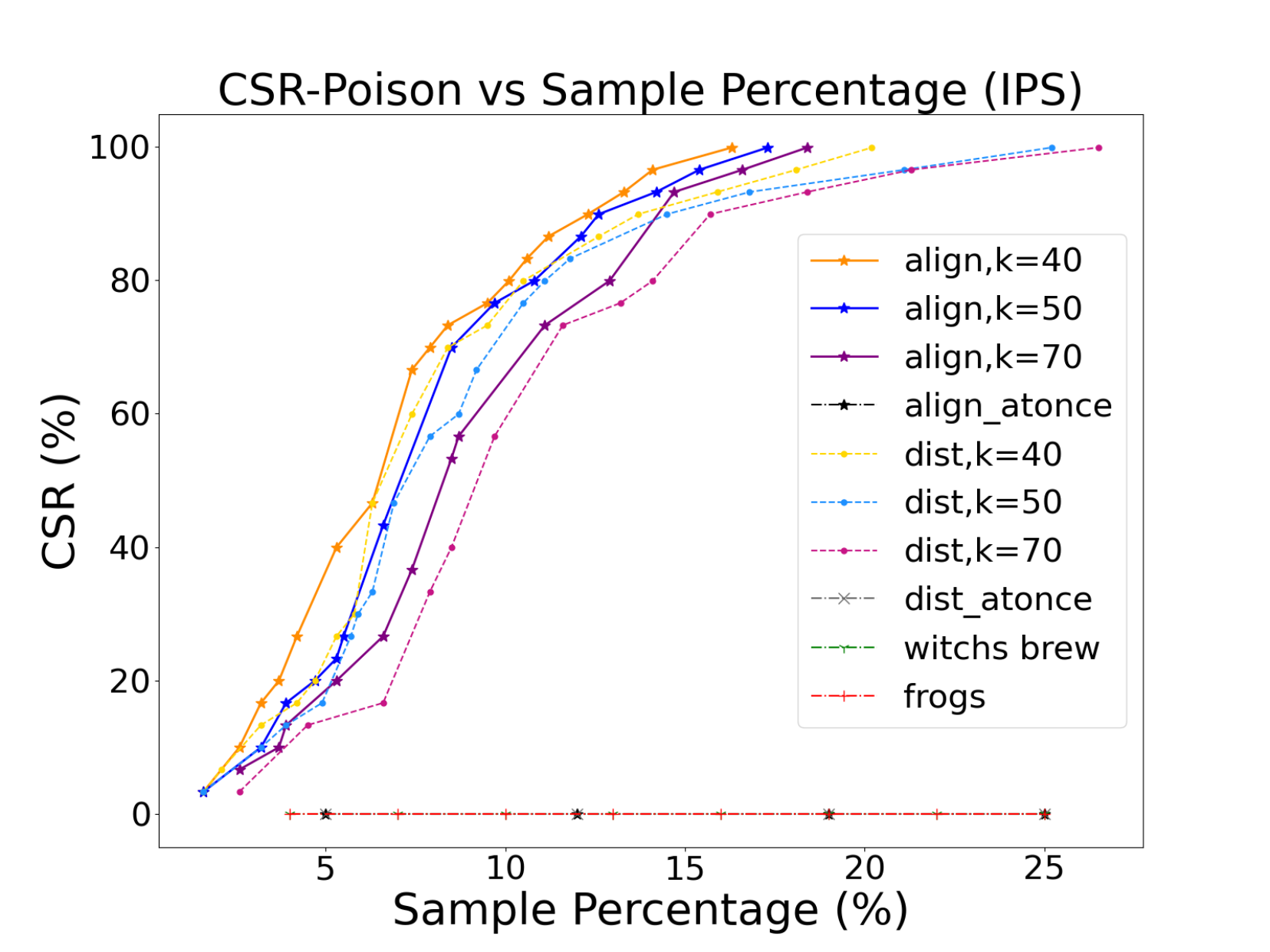}
\vspace{.3in}
\caption{
The CSR of Different Methods in Prediction Tampering Tasks with Varying Sample Percentage in MALWARE, EHR, and IPS.}
\label{fig:malware_poison}
\end{figure}
\section{Experiments}

In this section, we present the dataset used to evaluate the performance of the DMT method and demonstrate its effectiveness on both prediction improvement and tempering tasks.

\subsection{Dataset}
We use three discrete datasets from different application domains to evaluate the performance of our proposed DMT method, which are EHR, MALWARE and IPS. The details of them are introduced in Appendix 2.

\subsection{Implementation Details}
The implementation details of our experimental results are shown in Appendix 3.

\subsection{Evaluation Tasks and Metrics}

We evaluate the performance of DMT from two perspectives. 
One is to use DMT to improve the performance of a student model, i.e., correcting the wrongly classified samples. The second perspective is to allow the teacher to tamper student model's performance maliciously and share the same problem setting of poisoning attacks. It is worth noting that the performance improvement task requires the successful manipulation of the entire group of samples, rather than just one sample, as in the performance tampering tasks. In other words, 
the break condition in Algorithm \ref{alg:DMT} line 8 requires that all target samples in one group should be correctly classified. Therefore, the performance improvement tasks are more challenging than performance tampering tasks.

In our comparison of different methods, we focus on both computational efficiency and performance. Regarding the efficiency, we evaluate the average time consumption for each iteration, the average number of iterations required and the average percentage of samples needed over the entire dataset to achieve successful manipulation. Furthermore, we examine the change success rate (CSR), i.e., the number of correct classification of target samples as their desired class across all target samples. CSRs of different methods are compared under the constraint of the same maximum percentage of samples allowed to change. Finally, we consider the average decision logit of the last 5 training epochs as the output target logit to decide whether the target sample can be classified as the target class.

\subsection{Baseline Methods}
As there is no prior work of discrete machine teaching, we take a variant of our DMT as one baseline. There are two other models designed for the continuous domain. We adapt them to manipulate the discrete categorical features for performance improvement and prediction tampering tasks. 

\begin{itemize}
\item \textbf{At once} is a method based on our proposed method but with only one iteration. It can be considered as our method with a big step size $k$. It
only inserts the generated teacher dataset into the clean dataset so as to obtain the results at once.
\item \textbf{Frogs} \cite{shafahi2018poison} is a feature collision attack method, which proposed to move the base samples toward the target sample in the embedding space, rather than make the perturbation in the original feature space.
\item \textbf{Witch's brew} \cite{geiping2020witches} formulated the poisoning attack as a gradient matching problem and tackled the problem by aligning the target and poisoning gradients in the same direction.
\end{itemize}

Apart from the first baseline, the other two methods were proposed to solve the poisoning attack problem in the continuous domain. {Since the input data in the discrete domain are preprocessed as one hot vectors, we apply a rounding process to the samples output from these algorithms so that they can be converted into the discrete domain. 
For all baseline methods, we test their performances using different fixed numbers of manipulated samples to make a comparison.

\subsection{Prediction Improvement Results}

\begin{table*}[h]
    \centering
        \captionsetup{skip=1ex}
        \caption{Efficiency Evaluation   with Different Score Function on MALWARE and IPS Dataset for Improving Predictions}
        \captionsetup{skip=1ex}
    \begin{tabular}{c|c|c|c|c|c}
    \hline
    Dataset                 &Method                  &Step Size &Iteration &Time(min) &Sample Percent\\\hline
    \multirow{4}*{MALWARE}  &\multirow{2}*{DMT-Dist}  &k=10     &4.7       &\textbf{0.7}       &0.24\%        \\\cline{3-6}
                            &\multirow{2}*{}          &k=20     &4.0       &0.8       &0.36\%        \\\cline{2-6}
                            &\multirow{2}*{DMT-Align} &k=10     &2.6       &1.1       &\textbf{0.13}\%        \\\cline{3-6}
                            &\multirow{2}*{}          &k=20     &\textbf{2.3}       &1.5       &0.23\%        \\\hline
    \multirow{4}*{IPS}      &\multirow{2}*{DMT-Dist}  &k=40      &7.2       &\textbf{0.4}       &7.58\%        \\\cline{3-6}
                            &\multirow{2}*{}          &k=50      &7.2       &0.5       &9.47\%        \\\cline{2-6}
                            &\multirow{2}*{DMT-Align} &k=40      &6.2       &3.5       &\textbf{6.53}\%        \\\cline{3-6}
                            &\multirow{2}*{}          &k=50      &\textbf{5.4}       &4.6       &7.11\%        \\\hline
    \end{tabular}
    \label{tab:result_improve}
\end{table*}
\vspace{1ex}

The goal of a teacher is to correct the predictions of a group of wrongly classified samples. We report the results of our methods and baselines on MALWARE and IPS. The efficiency evaluation results are given in Table.\ref{tab:result_improve} and results of performance improvement are shown in Figure \ref{fig:malware_improve}.

In Figure \ref{fig:malware_improve}, the differences among evaluated methods may not be prominent. However, we can still observe some trends. Notably, the smaller step size leads to better improvement performances. Our proposed method using alignment score function outperforms the one using distance function. Besides, compared to baseline methods, our best setting, i.e., alignment function with small step size can achieve the highest CSR under the same sample percentage. As the sample percentage increases, the baseline methods exhibit similar improvement performance to the DMT-dist method, but the DMT-dist can eventually achieve 100\% CSR with a reasonable sample percentage limit.  

In terms of efficiency, from Table.\ref{tab:result_improve}, we can find that a larger step size results in fewer average iteration and required samples, but it also consumes more computing time. Compared with the distance method, the alignment score based method can reduce the number of iterations and generated samples, but it requires more time to achieve this reduction. 

\begin{table*}[h]
    \centering
        \captionsetup{skip=1ex}
        \caption{Efficiency  Evaluation with Different Score Function on MALWARE, IPS and EHR in Prediction Tampering Tasks}
        \captionsetup{skip=1ex}
    \begin{tabular}{c|c|c|c|c|c}
    \hline
    DataSet                &Method                 &Step Size &Iteration &Time(min) &Sample Percent\\\hline
    \multirow{6}*{MALWARE} &\multirow{3}*{DMT-Dist} &k=100     &14.2      &\textbf{8.8}      &7.12\%        \\\cline{3-6}
                           &\multirow{2}*{}          &k=200     &11.4      &10.2      &11.3\%        \\\cline{3-6}
                           &\multirow{2}*{}         &k=400     &11.1       &11.0      &22.28\%       \\\cline{2-6}
                          &\multirow{3}*{DMT-Align} &k=100     &11.4      &13.7      &\textbf{5.72}\%        \\\cline{3-6}
                          &\multirow{2}*{}          &k=200     &9.8       &23.9      &9.77\%        \\\cline{3-6}
                          &\multirow{2}*{}          &k=400     &\textbf{8.1}       &44.6      &16.27\%        \\\hline
    \multirow{6}*{IPS}    &\multirow{3}*{DMT-Dist}   &k=40      &9.8      &\textbf{1.2}       &8.47\%        \\\cline{3-6}
                          &\multirow{2}*{}          &k=50      &8.0       &1.6       &9.27\%        \\\cline{3-6}
                          &\multirow{2}*{}           &k=70      &6.2       &2.1       &10.50\%       \\\cline{2-6}
                          &\multirow{3}*{DMT-Align}   &k=40      &9.1       &3.1       &\textbf{8.15}\%        \\\cline{3-6}
                          &\multirow{2}*{}          &k=50      &7.3       &4.1      &8.43\%        \\\cline{3-6}
                          &\multirow{2}*{}          &k=70      &\textbf{5.7}       &5.7      &9.68\%        \\\hline                               
    \multirow{6}*{EHR}    &\multirow{3}*{DMT-Dist}    &k=10      &9.6      &\textbf{24.0}       &1.51\%        \\\cline{3-6}
                          &\multirow{2}*{}         &k=20      &7.1      &26.9       &2.21\%         \\\cline{3-6}
                          &\multirow{2}*{}         &k=40      &4.6      &32.7       &2.88\%         \\\cline{2-6}
                          &\multirow{3}*{DMT-Align}   &k=10      &7.8      &26.6       &\textbf{1.21}\%         \\\cline{3-6}
                          &\multirow{2}*{}         &k=20      &6.5      &30.5       &2.04\%         \\\cline{3-6}
                          &\multirow{2}*{}         &k=40      &\textbf{4.1}      &39.8       &2.58\%          \\\hline
                                   
    \end{tabular}
    \label{tab:result_poison}
\end{table*}
\vspace{1ex}

\subsection{Prediction Tampering Results}

We show the prediction tampering results of our machine teaching method in Table \ref{tab:result_poison}. 
The teacher plays a role of malicious attackers on tampering the prediction results on MALWARE, EHR and IPS dataset.
The performance of attacking is shown in 
Figure \ref{fig:malware_poison}. We evaluate our DMT using different score functions: distance and alignment score. For each DMT with a specific score function, the step size $k$ is set to 3 different values.  
The curves in Figure \ref{fig:malware_poison} illustrate the increase of CSR when more samples are introduced into $D_{teacher}$ in the iterative teaching process. All curves are compared at the same sample percentage. A small $k$ leads to a slow increase in $D_{teacher}$, while a large $k$ results in a faster increase. The baseline method \emph{at once} takes the entire set $D_{perturbed}$ at a given sample percentage into $D_{teacher}$ by one step, without running iteratively. 

The comparison results reveal several important findings. First, when the step size $k$ is smaller, a high CSR can be reached with a smaller percentage of samples. This indicates that a smaller step size can lead to more powerful manipulation. This is because a small $k$ value can accumulate the most powerful samples after enough iterations through interactive k-NN selection. It is also observed that our method outperforms all baseline methods. Moreover, comparing our DMT using distances and alignment score functions with the same step size, the alignment score function can result in greater CSRs. This aligns with our prediction that poisoned samples selected based on alignment score, using the gradients from the model are more powerful than ones selected based on the distance score. The alignment score function is more effective since it exploits the relation between manipulation samples and target samples.

Second, with the increase of step size, there is a noticeable rise in time consumption and the percentage of fake samples required to achieve successful manipulation. It indicates that no matter what the feature selection strategy is, a larger step size would lead to higher computational cost. Comparing the distance based method and the alignment score method, the latter can save more on sample perturbation and teaching iterations but consuming more time when crafting these samples. This trade-off between efficiency and performance should be considered when choosing a specific method for discrete machine teaching.

\section{Conclusion}

In this paper, we propose an iterative discrete machine teaching algorithm, involving teacher dataset construction and student model update. This is the first work investigating machine teaching problem in discrete domain without access to test data. We exploit score functions to select the most influential base samples from the dataset and then choose the most important features to change. 
Experimental results on three datasets demonstrate the effectiveness of our DMT in both performance improvement and performance tampering tasks, achieving strong and efficient manipulation with up to $100\%$ change success rate and minimal time and samples consumption. The results in this paper represent one step forward in the area of machine teaching on discrete data, and open up exciting avenues for future research, e.g., to investigate change success guarantees for given target samples.




\bibliographystyle{plainnat}
\bibliography{citation}

\begin{thebibliography}{49}
\providecommand{\natexlab}[1]{#1}
\providecommand{\url}[1]{\texttt{#1}}
\expandafter\ifx\csname urlstyle\endcsname\relax
  \providecommand{\doi}[1]{doi: #1}\else
  \providecommand{\doi}{doi: \begingroup \urlstyle{rm}\Url}\fi

\bibitem[Agarap(2018)]{agarap2018deep}
Abien~Fred Agarap.
\newblock Deep learning using rectified linear units (relu).
\newblock \emph{arXiv preprint arXiv:1803.08375}, 2018.

\bibitem[Aghakhani et~al.(2021)Aghakhani, Meng, Wang, Kruegel, and Vigna]{aghakhani2021bullseye}
Hojjat Aghakhani, Dongyu Meng, Yu-Xiang Wang, Christopher Kruegel, and Giovanni Vigna.
\newblock Bullseye polytope: A scalable clean-label poisoning attack with improved transferability.
\newblock In \emph{2021 IEEE European Symposium on Security and Privacy (EuroS\&P)}, pages 159--178. IEEE, 2021.

\bibitem[Akgun et~al.(2012)Akgun, Cakmak, Yoo, and Thomaz]{akgun2012trajectories}
Baris Akgun, Maya Cakmak, Jae~Wook Yoo, and Andrea~Lockerd Thomaz.
\newblock Trajectories and keyframes for kinesthetic teaching: A human-robot interaction perspective.
\newblock In \emph{Proceedings of the seventh annual ACM/IEEE international conference on Human-Robot Interaction}, pages 391--398, 2012.

\bibitem[Alfeld et~al.(2016)Alfeld, Zhu, and Barford]{alfeld2016data}
Scott Alfeld, Xiaojin Zhu, and Paul Barford.
\newblock Data poisoning attacks against autoregressive models.
\newblock In \emph{AAAI}, volume~30, 2016.

\bibitem[Bao et~al.(2021)Bao, Han, Zhou, Shen, and Zhang]{bao2021towards}
Hongyan Bao, Yufei Han, Yujun Zhou, Yun Shen, and Xiangliang Zhang.
\newblock Towards understanding the robustness against evasion attack on categorical data.
\newblock In \emph{ICLR}, 2021.

\bibitem[Bhadra and Hein(2015)]{bhadra2015correction}
Sahely Bhadra and Matthias Hein.
\newblock Correction of noisy labels via mutual consistency check.
\newblock \emph{Neurocomputing}, 160:\penalty0 34--52, 2015.

\bibitem[Cadamuro et~al.(2016)Cadamuro, Gilad-Bachrach, and Zhu]{cadamuro2016debugging}
Gabriel Cadamuro, Ran Gilad-Bachrach, and Xiaojin Zhu.
\newblock Debugging machine learning models.
\newblock In \emph{ICML Workshop on Reliable Machine Learning in the Wild}, volume 103, 2016.

\bibitem[Chan et~al.(2020)Chan, Tay, Ong, and Zhang]{chan2020poison}
Alvin Chan, Yi~Tay, Yew-Soon Ong, and Aston Zhang.
\newblock Poison attacks against text datasets with conditional adversarially regularized autoencoder.
\newblock \emph{arXiv preprint arXiv:2010.02684}, 2020.

\bibitem[Chen et~al.(2018)Chen, Singla, Mac~Aodha, Perona, and Yue]{chen2018understanding}
Yuxin Chen, Adish Singla, Oisin Mac~Aodha, Pietro Perona, and Yisong Yue.
\newblock Understanding the role of adaptivity in machine teaching: The case of version space learners.
\newblock \emph{Advances in Neural Information Processing Systems}, 31, 2018.

\bibitem[Cheng et~al.(2016)Cheng, Wang, Zhang, and Hu]{cheng2016risk}
Yu~Cheng, Fei Wang, Ping Zhang, and Jianying Hu.
\newblock Risk prediction with electronic health records: A deep learning approach.
\newblock In \emph{Proceedings of the 2016 SIAM international conference on data mining}, pages 432--440. SIAM, 2016.

\bibitem[Devlin et~al.(2018)Devlin, Chang, Lee, and Toutanova]{devlin2018bert}
Jacob Devlin, Ming-Wei Chang, Kenton Lee, and Kristina Toutanova.
\newblock Bert: Pre-training of deep bidirectional transformers for language understanding.
\newblock \emph{arXiv preprint arXiv:1810.04805}, 2018.

\bibitem[Doliwa et~al.(2014)Doliwa, Fan, Simon, and Zilles]{doliwa2014recursive}
Thorsten Doliwa, Gaojian Fan, Hans~Ulrich Simon, and Sandra Zilles.
\newblock Recursive teaching dimension, vc-dimension and sample compression.
\newblock \emph{The Journal of Machine Learning Research}, 15\penalty0 (1):\penalty0 3107--3131, 2014.

\bibitem[Fang et~al.(2020)Fang, Gong, and Liu]{fang2020influence}
Minghong Fang, Neil~Zhenqiang Gong, and Jia Liu.
\newblock Influence function based data poisoning attacks to top-n recommender systems.
\newblock In \emph{Proceedings of The Web Conference 2020}, pages 3019--3025, 2020.

\bibitem[Geiping et~al.(2020)Geiping, Fowl, Huang, Czaja, Taylor, Moeller, and Goldstein]{geiping2020witches}
Jonas Geiping, Liam Fowl, W~Ronny Huang, Wojciech Czaja, Gavin Taylor, Michael Moeller, and Tom Goldstein.
\newblock Witches' brew: Industrial scale data poisoning via gradient matching.
\newblock \emph{arXiv preprint arXiv:2009.02276}, 2020.

\bibitem[Goldman and Kearns(1995)]{goldman1995complexity}
Sally~A Goldman and Michael~J Kearns.
\newblock On the complexity of teaching.
\newblock \emph{Journal of Computer and System Sciences}, 50\penalty0 (1):\penalty0 20--31, 1995.

\bibitem[Goodfellow et~al.(2014)Goodfellow, Shlens, and Szegedy]{goodfellow2014explaining}
Ian~J Goodfellow, Jonathon Shlens, and Christian Szegedy.
\newblock Explaining and harnessing adversarial examples.
\newblock \emph{arXiv preprint arXiv:1412.6572}, 2014.

\bibitem[Guo and Liu(2020)]{guo2020practical}
Junfeng Guo and Cong Liu.
\newblock Practical poisoning attacks on neural networks.
\newblock In \emph{ECCV}, pages 142--158. Springer, 2020.

\bibitem[Guo et~al.(2022)Guo, Tondi, and Barni]{guo2022overview}
Wei Guo, Benedetta Tondi, and Mauro Barni.
\newblock An overview of backdoor attacks against deep neural networks and possible defences.
\newblock \emph{IEEE Open Journal of Signal Processing}, 2022.

\bibitem[Han and Zhang(2020)]{han2020robust}
Yufei Han and Xiangliang Zhang.
\newblock Robust federated learning via collaborative machine teaching.
\newblock In \emph{AAAI}, volume~34, pages 4075--4082, 2020.

\bibitem[He et~al.(2016)He, Zhang, Ren, and Sun]{he2016deep}
Kaiming He, Xiangyu Zhang, Shaoqing Ren, and Jian Sun.
\newblock Deep residual learning for image recognition.
\newblock In \emph{Proceedings of the IEEE conference on computer vision and pattern recognition}, pages 770--778, 2016.

\bibitem[Hochreiter and Schmidhuber(1997)]{hochreiter1997long}
Sepp Hochreiter and J{\"u}rgen Schmidhuber.
\newblock Long short-term memory.
\newblock \emph{Neural computation}, 9\penalty0 (8):\penalty0 1735--1780, 1997.

\bibitem[Jagielski et~al.(2021)Jagielski, Severi, Pousette~Harger, and Oprea]{jagielski2021subpopulation}
Matthew Jagielski, Giorgio Severi, Niklas Pousette~Harger, and Alina Oprea.
\newblock Subpopulation data poisoning attacks.
\newblock In \emph{Proceedings of the 2021 ACM SIGSAC Conference on Computer and Communications Security}, pages 3104--3122, 2021.

\bibitem[Kane et~al.(2017)Kane, Lovett, Moran, and Zhang]{kane2017active}
Daniel~M Kane, Shachar Lovett, Shay Moran, and Jiapeng Zhang.
\newblock Active classification with comparison queries.
\newblock In \emph{2017 IEEE 58th Annual Symposium on Foundations of Computer Science (FOCS)}, pages 355--366. IEEE, 2017.

\bibitem[Kingma and Ba(2014)]{kingma2014adam}
Diederik~P Kingma and Jimmy Ba.
\newblock Adam: A method for stochastic optimization.
\newblock \emph{arXiv preprint arXiv:1412.6980}, 2014.

\bibitem[Koh and Liang(2017)]{koh2017understanding}
Pang~Wei Koh and Percy Liang.
\newblock Understanding black-box predictions via influence functions.
\newblock In \emph{ICML}, pages 1885--1894. PMLR, 2017.

\bibitem[Koh et~al.(2022)Koh, Steinhardt, and Liang]{koh2022stronger}
Pang~Wei Koh, Jacob Steinhardt, and Percy Liang.
\newblock Stronger data poisoning attacks break data sanitization defenses.
\newblock \emph{Machine Learning}, 111\penalty0 (1):\penalty0 1--47, 2022.

\bibitem[Kouliaridis and Kambourakis(2021)]{kouliaridis2021comprehensive}
Vasileios Kouliaridis and Georgios Kambourakis.
\newblock A comprehensive survey on machine learning techniques for android malware detection.
\newblock \emph{Information}, 12\penalty0 (5):\penalty0 185, 2021.

\bibitem[Kurita et~al.(2020)Kurita, Michel, and Neubig]{kurita2020weight}
Keita Kurita, Paul Michel, and Graham Neubig.
\newblock Weight poisoning attacks on pre-trained models.
\newblock \emph{arXiv preprint arXiv:2004.06660}, 2020.

\bibitem[Lei et~al.(2018)Lei, Wu, Chen, Dimakis, Dhillon, and Witbrock]{lei2018discrete}
Qi~Lei, Lingfei Wu, Pin-Yu Chen, Alexandros~G Dimakis, Inderjit~S Dhillon, and Michael Witbrock.
\newblock Discrete attacks and submodular optimization with applications to text classification.
\newblock \emph{arXiv preprint arXiv:1812.00151}, 2018.

\bibitem[Liu et~al.(2016)Liu, Zhu, and Ohannessian]{liu2016teaching}
Ji~Liu, Xiaojin Zhu, and Hrag Ohannessian.
\newblock The teaching dimension of linear learners.
\newblock In \emph{International Conference on Machine Learning}, pages 117--126. PMLR, 2016.

\bibitem[Liu et~al.(2017)Liu, Dai, Humayun, Tay, Yu, Smith, Rehg, and Song]{liu2017iterative}
Weiyang Liu, Bo~Dai, Ahmad Humayun, Charlene Tay, Chen Yu, Linda~B Smith, James~M Rehg, and Le~Song.
\newblock Iterative machine teaching.
\newblock In \emph{ICML}, pages 2149--2158. PMLR, 2017.

\bibitem[Ma et~al.(2017)Ma, Chitta, Zhou, You, Sun, and Gao]{ma2017}
Fenglong Ma, Radha Chitta, Jing Zhou, Quanzeng You, Tong Sun, and Jing Gao.
\newblock Dipole: Diagnosis prediction in healthcare via attention-based bidirectional recurrent neural networks.
\newblock In \emph{KDD}, KDD '17, page 1903–1911, 2017.

\bibitem[Mei and Zhu(2015)]{mei2015using}
Shike Mei and Xiaojin Zhu.
\newblock Using machine teaching to identify optimal training-set attacks on machine learners.
\newblock In \emph{AAAI}, 2015.

\bibitem[Mu{\~n}oz-Gonz{\'a}lez et~al.(2019)Mu{\~n}oz-Gonz{\'a}lez, Pfitzner, Russo, Carnerero-Cano, and Lupu]{munoz2019poisoning}
Luis Mu{\~n}oz-Gonz{\'a}lez, Bjarne Pfitzner, Matteo Russo, Javier Carnerero-Cano, and Emil~C Lupu.
\newblock Poisoning attacks with generative adversarial nets.
\newblock \emph{arXiv preprint arXiv:1906.07773}, 2019.

\bibitem[Salem et~al.(2022)Salem, Wen, Backes, Ma, and Zhang]{salem2022dynamic}
Ahmed Salem, Rui Wen, Michael Backes, Shiqing Ma, and Yang Zhang.
\newblock Dynamic backdoor attacks against machine learning models.
\newblock In \emph{2022 IEEE 7th European Symposium on Security and Privacy (EuroS\&P)}, pages 703--718. IEEE, 2022.

\bibitem[Schuster et~al.(2020)Schuster, Schuster, Meri, and Shmatikov]{schuster2020humpty}
Roei Schuster, Tal Schuster, Yoav Meri, and Vitaly Shmatikov.
\newblock Humpty dumpty: Controlling word meanings via corpus poisoning.
\newblock In \emph{2020 IEEE Symposium on Security and Privacy (SP)}, pages 1295--1313. IEEE, 2020.

\bibitem[Shafahi et~al.(2018)Shafahi, Huang, Najibi, Suciu, Studer, Dumitras, and Goldstein]{shafahi2018poison}
Ali Shafahi, W~Ronny Huang, Mahyar Najibi, Octavian Suciu, Christoph Studer, Tudor Dumitras, and Tom Goldstein.
\newblock Poison frogs! targeted clean-label poisoning attacks on neural networks.
\newblock \emph{Advances in neural information processing systems}, 31, 2018.

\bibitem[Simard et~al.(2017)Simard, Amershi, Chickering, Pelton, Ghorashi, Meek, Ramos, Suh, Verwey, Wang, et~al.]{simard2017machine}
Patrice~Y Simard, Saleema Amershi, David~M Chickering, Alicia~Edelman Pelton, Soroush Ghorashi, Christopher Meek, Gonzalo Ramos, Jina Suh, Johan Verwey, Mo~Wang, et~al.
\newblock Machine teaching: A new paradigm for building machine learning systems.
\newblock \emph{arXiv preprint arXiv:1707.06742}, 2017.

\bibitem[Wallace et~al.(2020)Wallace, Zhao, Feng, and Singh]{wallace2020concealed}
Eric Wallace, Tony~Z Zhao, Shi Feng, and Sameer Singh.
\newblock Concealed data poisoning attacks on nlp models.
\newblock \emph{arXiv preprint arXiv:2010.12563}, 2020.

\bibitem[Wang et~al.(2021)Wang, Xu, Guzm{\'a}n, El-Kishky, Tang, Rubinstein, and Cohn]{wang2021putting}
Jun Wang, Chang Xu, Francisco Guzm{\'a}n, Ahmed El-Kishky, Yuqing Tang, Benjamin~IP Rubinstein, and Trevor Cohn.
\newblock Putting words into the system's mouth: A targeted attack on neural machine translation using monolingual data poisoning.
\newblock \emph{arXiv preprint arXiv:2107.05243}, 2021.

\bibitem[Wang et~al.(2020)Wang, Han, Bao, Shen, Ma, Li, and Zhang]{wang2020attackability}
Yutong Wang, Yufei Han, Hongyan Bao, Yun Shen, Fenglong Ma, Jin Li, and Xiangliang Zhang.
\newblock Attackability characterization of adversarial evasion attack on discrete data.
\newblock In \emph{KDD}, pages 1415--1425, 2020.

\bibitem[Yang et~al.(2017)Yang, Wu, Li, and Chen]{yang2017generative}
Chaofei Yang, Qing Wu, Hai Li, and Yiran Chen.
\newblock Generative poisoning attack method against neural networks.
\newblock \emph{arXiv preprint arXiv:1703.01340}, 2017.

\bibitem[Yang et~al.(2021)Yang, Li, Zhang, Ren, Sun, and He]{yang2021careful}
Wenkai Yang, Lei Li, Zhiyuan Zhang, Xuancheng Ren, Xu~Sun, and Bin He.
\newblock Be careful about poisoned word embeddings: Exploring the vulnerability of the embedding layers in nlp models.
\newblock \emph{arXiv preprint arXiv:2103.15543}, 2021.

\bibitem[Yang et~al.(2019)Yang, Dai, Yang, Carbonell, Salakhutdinov, and Le]{yang2019xlnet}
Zhilin Yang, Zihang Dai, Yiming Yang, Jaime Carbonell, Russ~R Salakhutdinov, and Quoc~V Le.
\newblock Xlnet: Generalized autoregressive pretraining for language understanding.
\newblock \emph{Advances in neural information processing systems}, 32, 2019.

\bibitem[Zhang et~al.(2018)Zhang, Zhu, and Wright]{zhang2018training}
Xuezhou Zhang, Xiaojin Zhu, and Stephen Wright.
\newblock Training set debugging using trusted items.
\newblock In \emph{AAAI}, volume~32, 2018.

\bibitem[Zhao et~al.(2017)Zhao, Dua, and Singh]{zhao2017generating}
Zhengli Zhao, Dheeru Dua, and Sameer Singh.
\newblock Generating natural adversarial examples.
\newblock \emph{arXiv preprint arXiv:1710.11342}, 2017.

\bibitem[Zhu(2013)]{zhu2013machine}
Jerry Zhu.
\newblock Machine teaching for bayesian learners in the exponential family.
\newblock \emph{Advances in Neural Information Processing Systems}, 26, 2013.

\bibitem[Zhu(2015)]{zhu2015machine}
Xiaojin Zhu.
\newblock Machine teaching: An inverse problem to machine learning and an approach toward optimal education.
\newblock In \emph{AAAI}, volume~29, 2015.

\bibitem[Zilles et~al.(2011)Zilles, Lange, Holte, Zinkevich, and Cesa-Bianchi]{zilles2011models}
Sandra Zilles, Steffen Lange, Robert Holte, Martin Zinkevich, and Nicol{\`o} Cesa-Bianchi.
\newblock Models of cooperative teaching and learning.
\newblock \emph{Journal of Machine Learning Research}, 12\penalty0 (2), 2011.

\end{thebibliography}
%

\end{document}